\documentclass{article}

\PassOptionsToPackage{numbers, compress}{natbib}

\usepackage[square,numbers]{natbib}
\usepackage[final]{neurips_2022}

\usepackage[utf8]{inputenc} 
\usepackage[T1]{fontenc}    
\usepackage{hyperref}       
\usepackage{url}            
\usepackage{booktabs}       
\usepackage{amsfonts}       
\usepackage{nicefrac}       
\usepackage{microtype}      
\usepackage{xcolor}         
\usepackage{graphicx}
\usepackage{amsmath}
\usepackage{multirow}
\usepackage{multicol}
\usepackage{comment}

\title{Physics Informed Symbolic Networks}

\author{%
  Ritam Majumdar \\
  TCS Research\\
  \texttt{ritam.majumdar@tcs.com} \\
  \And
  Vishal Jadhav\\
  TCS Research\\
  \texttt{vi.suja@tcs.com}\\
  \AND
  Anirudh Deodhar \\
  TCS Research\\\
  \texttt{anirudh.deodhar@tcs.com}\\
  \And
  Shirish Karande\\
  TCS Research\\
  \texttt{shirish.karande@tcs.com}\\
  \And
  Lovekesh Vig\\
  TCS Research\\
  \texttt{lovekesh.vig@tcs.com}\\
  \And
  Venkataramana Runkana\\
  TCS Research\\
  \texttt{venkat.runkana@tcs.com}\\
}

\begin{document}

\maketitle

\begin{abstract}
We introduce Physics Informed Symbolic Networks (PISN) which  utilize physics-informed loss to obtain a symbolic solution for a system of Partial Differential Equations (PDE). Given a context-free grammar to describe the language of symbolic expressions, we propose to use weighted sum as continuous approximation for selection of a production rule. We use this approximation to define multilayer symbolic networks. We consider Kovasznay flow (Navier-Stokes) and two-dimensional viscous Burger's equations to illustrate that PISN are able to provide a performance comparable to PINNs across various start-of-the-art advances: multiple outputs and governing equations, domain-decomposition, hypernetworks. Furthermore, we propose Physics-informed Neurosymbolic Networks (PINSN) which employ a multilayer perceptron (MLP) operator to model the residue of symbolic networks. PINSNs are observed to give 2-3 orders of performance gain over standard PINN.

\end{abstract}

\section{Introduction}
Physics Informed Neural Networks \cite{RAISSI} (PINNs) have gained popularity for numerically solving PDEs. However in the absence of an explicit inductive bias, MLPs can struggle to learn even simple functions in a sample efficient manner \cite{Periodic, NALU}, Appendix \ref{MLP_Extrapolation}. Meanwhile a symbolic form has the potential to exhibit better systematic generalization, with interpretability as a useful byproduct. Although symbolic regression has been studied extensively using a variety of algorithms, such as tree search \cite{AIfeymann, Aifeymann2, Neural-MCTS}, sequence generation \cite{SymbolicGPT, Robustfill}, sparse selection from a bank of primitive expressions \cite{SISSO} and genetic programming \cite{GP}, none of the existing works have explored the use of physics-informed loss (as done in PINNs) for obtaining a symbolic solution for a system of complex PDEs. We believe that potentially many of the above mentioned approaches can be extended to accommodate a physics-informed loss. However, our preliminary experimentation (\ref{SymRegComparison}) with \cite{Aifeymann2, SymbolicGPT} led to poor performance for complex examples of PDEs. Hence in this work we consider the problem of physics-informed symbolic regression and make three major contributions towards it. 

First, we propose Physics Informed Symbolic Networks (PISNs) that utilize differentiable programming and physics-informed loss for symbolic regression. We assume that the desired expression belongs to a language  defined by a context free grammar with [sin, exp, +, *] as symbolic operations. Inspired by recent works on differential synthesis of programs \cite{DSfPA, DARTS}, we approximate a selection of production rule by a weighted linear sum of the individual rules. This approximation is used to induce a multilayer symbolic network. We illustrate that PISNs provide a performance at par with PINNs, with Kovasznay flow (Navier Stokes) and two-dimensional viscous-Burger's equation as examples.



Second, we demonstrate that physics-informed training of the symbolic networks can be extended to recent advances in PINN, such as HyperNetworks\cite{HyperPINN} and Domain-decomposition \cite{cPINN, PHM} that have enabled applying PINNs to multiple outputs and governing equations, along with faster training.


Lastly, one cannot guarantee universal approximation even with larger number of compositions over an arbitrary set of operators. Therefore similar to \cite{DSfPA}, we allow the MLP operator to be a part of the grammar. Specifically, we capture the residual of the symbolic approximation with a MLP and show that a Physics Informed Neuro-Symbolic Network (PINSN) is able to provide 2-3 orders of magnitude performance gain over PINNs in terms of the maximum error.

\section{Methodology}

\subsection{Symbolic Networks}
\label{Symbolic_Networks}
We consider symbolic expressions that belong to a language defined by the context-free grammar, $\mathbb{G}$ \cite{CFG}, described by \eqref{CFG}, where x, y, t, c are the terminal symbols. We employ a continuous relaxation of $\mathbb{G}$ by taking a weighted sum over all the production rules as shown in \eqref{CCFG}. 
\begin{equation}\label{CFG}
\alpha::= \textbf{sin}\:\alpha_1\:|\:\textbf{exp}\:\alpha_1\:|\: \textbf{Add}\:\alpha_1\:\alpha_2\:|\: \textbf{Multiply}\:\alpha_1\:\alpha_2\:|\:x\:|\:y\:|\:t\:|\:c
\end{equation}
\begin{equation} \label{CCFG}
\alpha::= w_1\:\textbf{sin}(\alpha_1)+w_2\:\textbf{exp}(\alpha_2)+w_3(\alpha_3\textbf{+}\alpha_4)+w_4(\alpha_5\textbf{*}\alpha_6)+w_5x+w_6y+w_7t+w_81  
\end{equation}
Eqn. \eqref{CCFG} is used to induce a Symbolic Network of depth $d$+2, where $d$ is the total number of hidden layers. In the following equations, $\textbf{x}=[x,y,t,1]$ is the input, $\textbf{W}^0_k = [w^0_{k1},w^0_{k2},w^0_{k3},w^0_{k4}]$ are the weight vectors for the first hidden layer and $\textbf{W}^j_k = [w^j_{k1},w^j_{k2},w^j_{k3},w^j_{k4},w^j_{k5},w^j_{k6},w^j_{k7},w^j_{k8}]$ are the weight vectors for all the other hidden layers and $\textbf{W}^d$ are the weights used at the output layer. 
\begin{subequations}
\begin{align}
\textbf{u}_{pisn} &= \textbf{W}^d \cdot \textbf{h}^d\\
\textbf{h}^{j+1} & = [sin(\textbf{W}^{j}_1\cdot\textbf{h}^{j}),
    exp(\textbf{W}^{j}_2\cdot\textbf{h}^{j}),
    \textbf{W}^{j}_3\cdot\textbf{h}^{j}+\textbf{W}^{j}_4\cdot\textbf{h}^{j},
    \textbf{W}^{j}_5\cdot\textbf{h}^{j}*\textbf{W}^{j}_6\cdot\textbf{h}^{j},
    x,y,t]\\
\textbf{h}^1 &= [sin(\textbf{W}^{0}_1\cdot \textbf{x}),
    exp(\textbf{W}^{0}_2\cdot \textbf{x}),
    \textbf{W}^{0}_3\cdot \textbf{x}+\textbf{W}^{0}_4\cdot \textbf{x},
    \textbf{W}^{0}_5\cdot \textbf{x}*\textbf{W}^{0}_6\cdot \textbf{x},
    x,y,t]  
\end{align}
\end{subequations}
Eqn. \eqref{CCFG} could be used to induce a multilayer symbolic network similar to the differentiable program architectures proposed in \cite{DSfPA}. Such an approach would guarantee that the network is capable of exact representation of all the expressions with a depth proportional to the depth of the network. Nevertheless in such an architecture the number of parameters would grow exponentially with depth. On the contrary, the network defined by Eqn. (3) ensures that the number of parameters increases linearly with depth. However this does come at the cost of losing out on the power to exactly represent "wide" expressions. In spite of this, we have  observed that for the examples considered in this work, the approximations provided by our symbolic network are comparable with a symbolic network based upon \cite{DSfPA}, with neither approach providing a consistently better accuracy. Our approach is able to sometimes provide a advantage in terms of training time. Detailed investigation of this issue remains. 


\subsection{NeuroSymbolic Networks}
The expressive capacity of $\mathbb{G}$ can be improved by adding a MLP operator, $\textbf{F}(;\theta)$ as an additional parameterized terminal symbol to the grammar:
\begin{equation} \label{NsymN}
\alpha::= \textbf{sin}\:\alpha_1\:|\:\textbf{exp}\:\alpha_1\:|\: \textbf{Add}\:\alpha_1\:\alpha_2\:|\: \textbf{Multiply}\:\alpha_1\:\alpha_2\:|\:x\:|\:y\:|\:t\:|\:c\:|\:\textbf{F}(\textbf{x};\theta)
\end{equation}
Similar to symbolic networks we can make a continuous approximation of \eqref{NsymN} as well. However instead of adding a neural operator at all depths, for examples considered in this work, we observed substantial gains even by just approximating the residue of the symbolic networks. Therefore we define the neurosymbolic networks as follows. 
\begin{equation}
\label{PINSN_eqn}
\textbf{u}_{pinsn} = \textbf{u}_{pisn} + F(\textbf{x}; \theta_{res})
\end{equation}
\subsection{Hypernetworks}

We propose HyperPISN which consists of a hypernetwork $H_{pisn}$ which takes in the task parameterization $\lambda_h$ as input and outputs $\theta_{pisn}$ (the weights of the PISN network $M_{pisn}$). We consider examples where the initial conditions, boundary conditions and the coefficients of the PDE are all determined by single parameter $\lambda$. We use different values of $\lambda$ to sample tasks and train the hypernetwork, but show generalization to unseen value of $\lambda$. Since $\theta_{pisn}$ can be used to determine the final symbolic expression, HyperPISN can produce symbolic expressions for parameterized PDEs . 
\begin{equation}
\theta_{pisn} = H_{pisn} (\lambda_h; \theta_s),\:\:\:\textbf{u}_{pisn} = M_{pisn}(\textbf{x};\theta_{pisn})\\    
\end{equation}
We extend PINSNs to HyperPINSNs whose training follows a two step approach. In first step, $H_{pisn}$ (HyperPISN) is trained for a set of task parameters $\lambda$. In second step, HyperPISN weights are frozen and $H_{res}$ (hypernetwork of MLP operator $F$)  is trained to predict $\theta_{res}$ (weights of $F$).
\begin{equation}
\theta_{pisn} = H_{pisn} (\lambda_h; \theta_s),\:\:\:\theta_{res} = H_{res}(\lambda_h; \theta_h),\:\:\:\textbf{u}_{pinsn} = M_{pisn} (\textbf{x}; \theta_{pisn}) + F(\textbf{x};\theta_{res})\\    
\end{equation}
\section{Experiments}
\subsection{Kovasznay flow (Navier Stokes)}
Kovasznay flow is a laminar flow governed by two-dimensional steady state Navier-Stokes equation. It's governing equations and true analytical solutions are mentioned in Table \ref{NS:PDE Information}. We obtain the boundary conditions from the exact solutions mentioned in Table \ref{NS:PDE Information}. The governing equations and boundary conditions are parameterizeed by the Reynolds number Re $\in$[100,1000]. We train the hypernetwork architectures by using 10 equidistant train-tasks. 
We elaborate on the training process, hyperparameter details and the mean-L2-Error results in Appendix \ref{NS-Kovasznay-details}.

\begin{table}
\centering
\caption{Kovasznay Flow: Pointwise-Max-L2-Error viz-a-viz exact analytical solution}
\begin{tabular}{c c c c c c c c}
&&\multicolumn{3}{c}{Hypernetwork}&\multicolumn{3}{c}{Vanilla}\\
\hline
Re&Architecture&X-velocity&Y-velocity&Pressure&X-velocity&Y-velocity&Pressure\\
\hline
\multirow{3}{*}{125}&PINN&1.76e-5&4.95e-5&1.26e-5&4.93e-6&2.41e-5&4.23e-6\\
&PISN&6.88e-4&7.53e-4&2.00e-4&5.89e-5&5.49e-5&5.87e-5\\
&PINSN&\textbf{1.04e-6}&\textbf{1.27e-6}&\textbf{4.54e-6}&\underline{\textbf{7.79e-7}}&\underline{\textbf{6.06e-7}}&\underline{\textbf{2.44e-6}}\\
\hline
\multirow{3}{*}{375}&PINN&2.36e-5&2.27e-6&1.74e-5&1.19e-5&6.49e-6&1.33e-5\\
&PISN&2.86e-4&5.73e-5&4.58e-5&5.58e-5&1.26e-5&1.86e-5\\
&PINSN&\underline{\textbf{4.05e-7}}&\underline{\textbf{9.62e-8}}&\underline{\textbf{1.15e-7}}&\textbf{1.85e-6}&\textbf{4.05e-7}&\textbf{3.34e-7}\\
\hline
\multirow{3}{*}{475}&PINN&5.21e-5&5.48e-5&1.96e-5&7.72e-5&1.42e-6&6.34e-6\\
&PISN&2.33e-4&1.77e-5&1.75e-5&1.54e-5&4.09e-5&9.01e-5\\
&PINSN&\textbf{4.63e-7}&\underline{\textbf{1.49e-7}}&\textbf{1.35e-7}&\underline{\textbf{2.35e-7}}&\textbf{1.87e-7}&\underline{\textbf{3.34e-8}}\\
\hline
\multirow{3}{*}{725}&PINN&8.83e-6&5.10e-6&3.89e-6&9.39e-6&7.14e-6&4.22e-6\\
&PISN&3.41e-5&1.31e-5&2.02e-5&5.87e-5&4.31e-5&1.09e-5\\
&PINSN&\textbf{2.59e-7}&\textbf{3.34e-7}&\textbf{8.12e-7}&\underline{\textbf{4.29e-6}}&\underline{\textbf{1.87e-7}}&\underline{\textbf{3.34e-8}}\\
\hline
\multirow{3}{*}{975}&PINN&6.32e-5&2.37e-6&5.42e-5&3.54e-5&1.53e-6&4.37e-6\\
&PISN&1.19e-4&1.06e-5&1.09e-4&1.35e-4&4.19e-5&1.37e-4\\
&PINSN&\textbf{4.63e-6}&\underline{\textbf{4.14e-8}}&\textbf{1.91e-7}&\underline{\textbf{4.29e-6}}&\textbf{3.45e-7}&\underline{\textbf{6.60e-8}}\\
\hline
\end{tabular}
\vspace{1pt}
\label{tab:Navier-Stokes-Hypernetwork-Max}
\end{table}
\subsection{Two-dimensional coupled viscous Burger's equation}
\label{section:Coupled-Burgers}
The two-dimensional coupled viscous Burger's equation is used to model various physical phenomena like shock-wave propogation, turbulence, sedimentation, etc, whose governing equations and true analytical solutions are defined in Table \ref{2D-Burger's:PDE Information}. We consider the range of viscosity coefficient $\nu\in$[1e-3,1e-2] as parameter $\lambda$ for governing equations. 
In order to capture the dynamics in every sub-domain, we provide sub-domainwise comparative results for one test-task in Appendix \ref{2D-Coupled-Burgers-details} and figures \ref{fig:Burger's-Vanilla-u},\ref{fig:Burger's-Vanilla-v},\ref{fig:Burger's-Hyper-u},\ref{fig:Burger's-Hyper-v}. Additional information regarding pointwise mean-L2-error results, training schedule and hyperparameter details are mentioned in Appendix \ref{2D-Coupled-Burgers-details}. 

\begin{table}
\centering
\caption{2D Burger's: Pointwise-Max-L2-Error viz-a-viz exact analytical solution}
\begin{tabular}{c c c c c c}
&&\multicolumn{2}{c}{Hypernetwork}&\multicolumn{2}{c}{Vanilla}\\
\hline
$\nu$ (1e-3)&Architecture&X-velocity&Y-velocity&X-velocity&Y-velocity\\
\hline
\multirow{3}{*}{2.2}&PINN&7.65e-4&5.96e-3&7.55e-4&3.11e-3\\
&PISN&7.61e-3&6.98e-3&6.91e-3&3.96e-3\\
&PINSN&\underline{\textbf{1.95e-5}}&\underline{\textbf{2.21e-4}}&\underline{\textbf{3.14e-5}}&\underline{\textbf{3.96e-5}}\\
\hline
\multirow{3}{*}{4.3}&PINN&8.82e-4&2.18e-2&5.85e-4&4.37e-2\\
&PISN&1.33e-3&1.19e-2&3.97e-2&1.93e-2\\
&PINSN&\underline{\textbf{1.45e-4}}&\underline{\textbf{1.67e-4}}&\underline{\textbf{9.23e-5}}&\underline{\textbf{1.69e-3}}\\
\hline
\multirow{3}{*}{5.8}&PINN&3.38e-4&3.71e-2&3.28e-4&3.70e-2\\
&PISN&4.99e-3&7.05e-2&4.98e-3&7.04e-2\\
&PINSN&\underline{\textbf{4.17e-5}}&\underline{\textbf{7.25e-3}}&\underline{\textbf{3.61e-5}}&\underline{\textbf{7.21e-3}}\\
\hline
\multirow{3}{*}{7.5}&PINN&8.45e-3&8.22e-3&8.44e-3&8.21e-2\\
&PISN&2.73e-2&4.36e-2&2.73e-2&4.35e-2\\
&PINSN&\underline{\textbf{2.94e-5}}&\underline{\textbf{4.05e-3}}&\underline{\textbf{2.73e-4}}&\underline{\textbf{4.07e-4}}\\
\hline
\multirow{3}{*}{9.3}&PINN&3.93e-3&8.71e-3&3.92e-3&8.65e-3\\
&PISN&4.65e-3&2.56e-2&4.64e-3&2.55e-2\\
&PINSN&\underline{\textbf{2.25e-4}}&\underline{\textbf{7.57e-4}}&\underline{\textbf{1.92e-4}}&\underline{\textbf{6.46e-4}}\\
\hline
\end{tabular}
\vspace{1pt}
\label{tab:2D-Burger's-2-output-Domain_Hypernetwork-Max}
\end{table}


\section{Observations}

 We compare PISN with state-of-the-art Symbolic Regression methods in Appendix \ref{SymRegComparison}. From table \ref{SymRegCompare}, we observe AI-Feynman \cite{Aifeymann2} produces exact symbolic expressions for simpler examples, as the simplifying properties are easier to capture, but struggles for complex examples. SymbolicGPT \cite{SymbolicGPT} on the other hand fails to capture symbolic expressions even for simpler PDE setups without any noisy inputs. Multivariate Symbolic Function Learner (MSFL) \cite{Panju} performs better than SymbolicGPT and provides reasonably accurate solutions for simpler PDEs setups, however they fail to scale up to complex examples, as they are constrained by numerical approximations. 
 We observe, our PISN architecture outperforms all the benchmark methods for complex examples. We compare PISN to alternative symbolic differentiable program architecture methods in Table \ref{DiferentiableSymRegCompare}. We observe, PISNs outperform MSFL+Autograd on all tasks, and on an average, is 36.5 times better on the ratio of max-error of PISN and MSFL+Autograd. PISNs also have 1.2 times lower error than differentiable program architecture method proposed in \cite{DSfPA}, despite having lower number of parameters. 
 
 Tables \ref{tab:Navier-Stokes-Hypernetwork-Max} and \ref{tab:2D-Burger's-2-output-Domain_Hypernetwork-Max} shows the results for Kovasznay flow and coupled-Burger's equations respectively. We provide one illustrative example for Navier Stokes and it's corresponding error maps in Figures \ref{fig:NS} and \ref{fig:Error_NS}. For all examples across both the experiments, the difference of errors between a PISN architecture and it's corresponding PINN architecture is less than that of an order. This same observation holds for HyperPISNs, Domain-decomposed-PISNs, and Domain-decomposed-HyperPISNs, which implies recent-advances in PINNs can be combined with PISNs. 

When we compare the performance of HyperPISN with PISN we observe ratio of maximum error is on average less than 2 for the 2D Burger's equations and less than 4.1 for Kovasznay Flow equations. The increase in ratio of error in case of Kovasznay Flow can be largely attributed to case of $Re$ equal to 125 where the ratio of error averaged across the 3 outputs is ~9.6, which is still within an order of magnitude. We can therefore say that HyperPISNs are hyper symbolic solvers that can generalize across complex parameterized PDEs without the need for training on every value of $\lambda$. Its not straightforward to conceive such a generalization for other symbolic solvers.  

We observe PINSN outperform PINN and PISN by 2-3 orders of magnitude across all tasks, and this observation further extends to HyperPINSN, Domain-decomposed-PINSN, and Domain-decomposed-HyperPINSN. This implies adding a MLP operator to the grammar is highly advantageous, as that MLP operator is capable of capturing the residues of underlying symbolic expression. We provide an illustrative example for 1 Burger's equation test-case in Figures \ref{fig:Burger's-Vanilla-u}, \ref{fig:Burger's-Vanilla-v}, \ref{fig:Burger's-Hyper-u}, \ref{fig:Burger's-Hyper-v}. We notice, for every subdomain, PINSNs capture the residues from PISNs and have lower error magnitudes than PINNs from the error heat-maps.

\section{Limitations and Future Work}
\label{Limitations}

Our current setup for PISN suffers from training-collapse when we use operators which drastically change the symbolic expression search space e.g. division, logarithm, which restricts the choice of operators. This leads to our current expressions being long and cumbersome. For better interpretability of mathematical expressions, crisper analytical solutions are necessary. Hence, in future, we seek to explore techniques that incorporate additional operators.


\section{Broader Impact}

Modeling of a real world phenomenon in terms of differential equations is often an iterative process.
In this context, symbolic solutions not only lend themselves to increased interpretability but also allow for a human in the loop discovery of solutions of PDEs. We imagine that domain experts and scientist could often guess some approximate behavior of the solution. This can be used to constrain or initialize the weights of the symbolic network.

\bibliography{neurips_2022.bib}

\appendix

\section{Appendix}

Contents of appendix
\begin{itemize}
    \item \textbf{A.1} Preliminaries
    \subitem {\textbf{A.1.1}} Physics-Informed Loss
    \subitem {\textbf{A.1.2}} HyperPINN
    \item \textbf{A.2} Extrapolation failure of MLPs
    \item \textbf{A.3} PISN Vs Symbolic Regression Solvers
    \item \textbf{A.4} Additional Results and Training details
        \subitem \textbf{A.4.1} Navier Stokes: Kovasznay Flow
        \subitem \textbf{A.4.2} Two-dimensional coupled Burger's equation
    \item \textbf{A.5} Additional Differential Equations
        \subitem \textbf{A.5.1} Telegraph Equations
    \subitem \textbf{A.5.2} Two-dimensional Burgers: conservation equation
    \item \textbf{A.6} Sample Mathematical Expressions generated for experiments in section A.3
    \item \textbf{A.7} Impact of L-BFGS Optimizer and increasing the depth of Symbolic Network
\end{itemize}

\subsection{Preliminaries}
\subsubsection{Physics-Informed Loss}

We consider Partial Differential Equations of the form 
\begin{equation}
N[x,t,u(x,t)]=0, t\in[0,T], x\in\Omega 
\end{equation}
Where $N$ is a non-linear differential operator consisting of space and time derivatives, x and t refer to the spatial and temporal variables respectively. All the three architectures, PINN, PISN and PINSN employ a physics-informed loss to learn their respective parameters which is defined as follows:
\begin{equation}
L(\theta) = L_{Physics}(\theta) + L_{IDC}(\theta) + L_{INC}(\theta) + L_{BC}(\theta)\\   
\end{equation} where, 
\begin{subequations}
\begin{align}
L_{Physics}(\theta) &=  \sum_{x_c,t_c \in C}
N(u(x_c,t_c), x_c, t_c, \frac{du(x_c,t_c)}{dt}, \frac{du(x_c,t_c)}{dx}, \text{...})^2\\
L_{IDC}(\theta) &= \sum_{x_i \in IDC} (u(x_i,0)-f(x_i))^2\\
L_{INC}(\theta) &= \sum_{x_i \in INC} (\frac{\partial u(x_i,0)}{\partial t}-g(x_i))^2\\
L_{BC}(\theta) &= \sum_{x_{bc},t_{i} \in BC} (u(x_{bc},t_i)-h(x_{bc,t_i}))^2 
\end{align}
\end{subequations}
C represents the set of collocation points on which the Physics-loss is calculated. IDC, INC, BC represents the set of points at which Initial Dirichlet Conditions, Initial Neumann Conditions and Boundary Conditions are evaluated. $\lambda_i$, where i $\in$ {1,2,3,4} are scalers which are user-defined hyperparameters used to control the loss function. All the three networks, PINN, PISN and PINSN employ physics-informed loss to train their respective architectures. In case of PINN, $\theta$ represents the weights of the neural network, whereas for PISN and PINSN, $\theta$ represents the weights of the continuous approximation of rules defined by context-free grammar.

\subsubsection{HyperPINN}

Fillipe et al. in \cite{HyperPINN} introduced HyperPINNs to solve for parameterized differential equations (Lorenz System and one-dimensional Burger's equation). 
\begin{equation}
\theta_m = H_{pinn} (\lambda, \theta_h),  \textbf{u}_{pinn}(x,t) = M_{pinn}(x,t,\theta_m)
\end{equation}
HyperPINN consists of two components, a hypernetwork $H_{pinn}$ and a main-network $M_{pinn}$. $H_{pinn}$ takess in task parameterization $\lambda_h$ as input and outputs the weights of the main-network $\theta_m$. $\theta_m$ is directly supplied to the main-network for evaluating the solutions of the differential equation $\textbf{u}_{pinn}$. We use different values of $\lambda$ to sample tasks and train the hypernetwork, but show generalization to unseen value of $\lambda$.  

\subsection{Extrapolation Failure of MLPs}
\label{sec:Extrapolation}

In this section, we illustrate the extrapolation properties of MLPs for various well-known primitive functions. We consider a MLP with 6 hidden layers, 20 neurons each, considering 100,000 training points within the domain. Training domain extents for each function are - linear: [-1,1], exp:[1,3], log:[4,8], sin:[$-5\pi/8$,$5\pi/8$].
\begin{figure}[ht]
    \centering
    \caption{Extrapolation failure of MLPs for well-known primitive functions}
    \includegraphics[width=\textwidth]{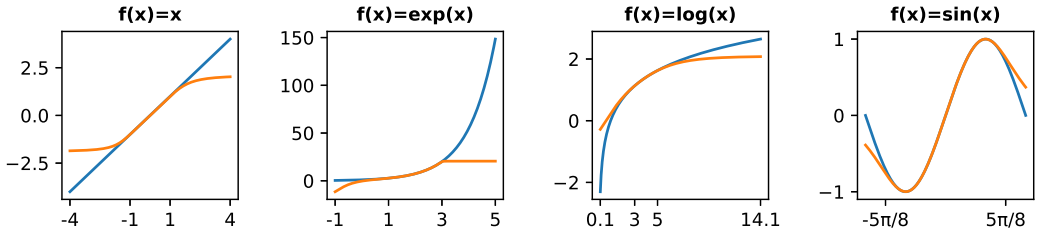}
    \label{MLP_Extrapolation}
\end{figure}
Blue and orange lines in Figure \ref{MLP_Extrapolation} denote the true function and the extrapolation by MLPs respectively. While MLPs have excellent interpolation properties, they are astonishingly poor extrapolators.

\subsection{PISN Vs Symbolic Regression Solvers}
\label{SymRegComparison}
In this section, we compare PISN to existing state-of-the-art Symbolic Regression solvers. AI-Feynman \cite{Aifeymann2} comprises simplifying properties of the functions of interest such as symmetry, compositionality and separability among others, to discover the symbolic expressions. SymbolicGPT \cite{SymbolicGPT} is a generative-transformer based language model to generate mathematical expressions based on a given set of input-output values. Panju et al. in \cite{Panju} developed a Multivariate Symbolic Function Learner (MSFL) to generate parse-trees, which can be used to extract mathematical expressions. While their Taylor-series approximations provide reasonably accurate approximations for lower order derivatives, they deteriorate for higher order and multivariate derivatives. On the other hand, Automatic differentiation \cite{Autograd} provides accurate derivatives even for higher order and multivariate derivatives. Hence, for our PISN architectures, we use automatic differentiation to approximate the physics-informed loss. Table \ref{Ablation_PDE_information} shows the PDE problem setups for the additional experiments taken from \cite{Panju} (used to benchmark MSFL). 

\begin{table}
    \centering
    \caption{PDE information for ablation experiments}
	\begin{tabular}{ c c c c c c}
      & B.C. & \(u(x,0)\) & \(u_t(x,0)\) & True Solution \\
     \hline
     \rule{0pt}{2ex} Wave: 
     & \(u(0,t)=0\)
     & \multirow{2}{*}{0}
     & \multirow{2}{*}{\(sin(x)\)}
     & \multirow{2}{*}{\(sin(x)sin(t)\)} \\
     \(u_{tt}=u_{xx}\)&\(u(\pi,t)=0\) &&&\\
     \hline
     \rule{0pt}{2ex} Heat:
     & \(u(0,t)=0\)
     & \multirow{2}{*}{\(sin(x)\)} 
     & \multirow{2}{*}{-}
     & \multirow{2}{*}{\(sin(x)e^{-t}\)}\\
     \(u_{t}=u_{xx}\)&\(u(\pi,t)=0\)&&&\\
     \hline
     \rule{0pt}{2ex} Fokker-Plank-1:
     & \multirow{2}{*}{-}
     & \multirow{2}{*}{\(x\)}
     & \multirow{2}{*}{-}
     & \multirow{2}{*}{\((x+t)\)}\\
     \(u_{t}=u_{x}+u_{xx}\)&& &&\\
     \hline
     \rule{0pt}{2ex} Fokker-Plank-2:
     & \multirow{4}{*}{-} 
     & \multirow{4}{*}{\(x\)}
     & \multirow{4}{*}{-}
     & \multirow{4}{*}{\(xe^t\)}\\
     \(u_{t}=p_{x}+q_{xx}\)&&&&\\
     \(p(x,t)=u(x,t)*x\)&&&&\\
     \(q(x,t)=u(x,t)*x^2\)&&&&\\
     \hline
     \rule{0pt}{2ex} Fokker-Plank-3:
     & \multirow{4}{*}{-} 
     & \multirow{4}{*}{\(x+1\)}
     & \multirow{4}{*}{-}
     & \multirow{4}{*}{\((x+1)e^t\)}\\
     \(u_{t}=p_{x}+q_{xx}\)&&&&\\
     \(p(x,t)=-u(x,t)*(x+1)\)&&&&\\
     \(q(x,t)=u(x,t)*x^2*e^t\)&&&&\\
     \hline
    \end{tabular}
    \vspace{1pt}
    \label{Ablation_PDE_information}
\end{table}

\begin{table}
\caption{PISN comparison with existing benchmarks}
\label{SymRegCompare}
\centering
\begin{tabular}{c c c c c c c}
&AI-FN&AI-FN&SGPT&SGPT&MSFL&PISN\\
&&(PINN)&&(PINN)&&\\
\hline
\vspace{1pt}
Wave&\textbf{Exact}&3.52e-7&1.24&1.38&1.35e-4&2.41e-7\\
\hline
Heat&\textbf{Exact}&2.98e-9&2.18&8.41&1.18e-4&2.99e-7\\
\hline
FP-1&\textbf{Exact}&1.68e-8&7.70e-2&6.16&1.26e-6&1.09e-9\\
\hline
FP-2&\textbf{Exact}&2.58e-7&6.36&2.85&1.74e-4&1.14e-7\\
\hline
FP-3&\textbf{Exact}&4.26e-6&6.35&9.18&8.86e-3&7.71e-7\\
\hline
NS (u)&0.21&5.71&0.91&0.84&1.29e-2&\textbf{1.03e-6}\\
NS (v)&1.00&1.70e+3&3.23&7.77&2.45e-2&\textbf{3.72e-6}\\
NS (P)&\textbf{5.43e-9}&6.02e-4&2.96&12.24&7.17e-2&5.01e-6\\
\hline
Burger's (u)&1.07e+3&1.52e+3&2.04e+3&4.09e+3&1.19e-1&\textbf{1.76e-4}\\
Burger's (v)&1.86e+3&2.45e+3&1.53e+3&3.90e+3&5.82e-2&\textbf{1.77e-4}\\
\hline
\end{tabular}
\end{table}

\begin{table}
\caption{Comparison of Symbolic differentiable program architecture methods}
\label{DiferentiableSymRegCompare}
\centering
\begin{tabular}{c c c c}
&MSFL&\multirow{2}{*}{PISN}&Program\\
&(Autograd)&&Architecture in \cite{DSfPA}\\
\hline
\vspace{1pt}
Wave&3.52e-6&2.41e-7&\textbf{1.59e-7}\\
\hline
Heat&8.84e-6&2.99e-7&\textbf{2.26e-7}\\
\hline
FP-1&5.37e-8&\textbf{1.09e-9}&2.84e-9\\
\hline
FP-2&4.51e-6&\textbf{1.14e-7}&1.58e-7\\
\hline
FP-3&4.24e-5&7.71e-7&\textbf{6.96e-7}\\
\hline
NS (u)&9.65e-5&\textbf{1.03e-6}&2.05e-6\\
NS (v)&2.91e-4&3.72e-6&\textbf{1.95e-6}\\
NS (P)&1.95e-5&\textbf{5.01e-6}&5.53e-6\\
\hline
Burger's (u)&9.36e-4&\textbf{1.76e-4}&1.91e-4\\
Burger's (v)&5.57e-4&\textbf{1.77e-4}&1.85e-4\\
\hline
\end{tabular}
\end{table}

\textbf{Experiment Design:} SymbolicGPT and AI-Feynman seek to fit mathematical expressions, given a set of input-output pairs. Since PDEs don't have labelled data, we use a PINN to generate input-output pairs over the entire domain. Since the input-output pairs are generated from outputs of a MLP, they are noisy. We then pass these data points as input to AI-Feynman and pre-trained SymbolicGPT models to extract the symbolic expression. As an additional benchmark, we also consider noiseless samples by generating the input-output pairs directly from the ground-truth analytical expression. Table \ref{SymRegCompare} consists the pointwise-mean-L2-error results, wherein NS refers to 2D Navier-Stokes: Kovasznay flow result for Reynolds number Re=475, and Burger's refers to two-dimensional coupled viscous Burger's equation for $\nu=4.3e-3$. The true analytical solutions for the PDEs are referred to from tables \ref{Ablation_PDE_information}, \ref{NS:PDE Information} and \ref{2D-Burger's:PDE Information}. We display the expressions generated by AI-Feynman and SymbolicGPT, and for 1 Navier-Stokes test-case in Appendix \ref{Expressions}. We use publicly available AI-Feynman and SymbolicGPT pretrained-models for conducting the experiments in this section. 

Table \ref{DiferentiableSymRegCompare} shows the comparison between differentiable symbolic regression architectures. We note, performance of MSFL architectures improves by 2-3 orders of magnitude when physics-loss is computed using automatic differentiation as against a Taylor series approximation. Differentiable programmable architectures refers to \cite{DSfPA}, which has a higher representation capacity as \# Parameters increase exponentially with depth. We compare the ratio of mean-errors of PISN with MSFL+Autograd, and observe that on an average, PISNs are 36.5 times better than MSFL+Autograd, and PISNs superiority is attributed by it's significantly
better performance in complex examples. Furthermore, the wave equation has a true solution of $sin(x)sin(t)$, which our PISN cannot exactly represent. Yet, our PISN architecture has a very low error of 2.41e-7, indicating the possibility to generate excellent approximations to complex PDEs which have no analytical solutions.

\subsection{Additional Results and Training details}
\label{sec:Hyperparameter-details}

\subsubsection{Kovasznay Flow (Navier Stokes)}
\label{NS-Kovasznay-details}

Table \ref{NS:PDE Information} represents the differential equations and true analytical solutions of Kovasznay flow (Navier Stokes). Here, the first equation represents the conservation of mass balance, while the second and third equations represent momentum transfer equations in X and Y dimension respectively. We consider a 101\textbf{x}101 equally spaced grid domain to represent the XY plane, where $X,Y\in$[-0.5,1.0]\textbf{X}[-0.5,1.5]. We consider 2601 collocation points and 320 boundary condition points, with 80 points for each face of the grid.  We consider 10 train-tasks equally spread out across the task parameterization space, 5 validation tasks to avoid meta-overfit and 5 test tasks for evaluation. We train all experiments for 125k epochs. Remaining hyperparameter details are mentioned in Table \ref{tab:NS-Hyperparameters}.

\begin{table}
  \caption{Navier-Stokes: Kovasznay Flow PDE Information}
  \label{NS:PDE Information}
  \centering
  \begin{tabular}{c c}
  \hline
  Governing Differential Equations & True Analytical solutions\\
  \hline
  \(u_{x} + v_{y} = 0\)&\(u(x,y) = 1-e^{\lambda x}cos(2\pi y)\)\\
  \(u*u_{x} + v*u_{y} = -p_{x} + (u_{xx} + u_{yy})/Re\)&\(v(x,y) = \lambda e^{\lambda x}sin(2\pi y)/2 \pi\)\\
  \(u*v_{x} + v*v_{y} = -p_{y} + (v_{xx} + v_{yy})/Re\)&\(p(x,y) = (1-e^{2 \lambda x})/2\)\\
  \hline
  \end{tabular}
\end{table}

\begin{table}
\centering
\caption{Kovasznay Flow: Pointwise-Mean-L2-error viz-a-viz true analytical solution}
\begin{tabular}{c c c c c c c c}
&&\multicolumn{3}{c}{Hypernetwork}&\multicolumn{3}{c}{Vanilla}\\
\hline
Re&Architecture&X-velocity&Y-velocity&Pressure&X-velocity&Y-velocity&Pressure\\
\hline
\multirow{3}{*}{125}&PINN&1.76e-6&6.95e-7&1.15e-6&4.11e-7&2.01e-6&\underline{\textbf{3.02e-7}}\\
&PISN&5.30e-4&7.53e-5&2.00e-5&5.89e-5&3.92e-6&6.52e-6\\
&PINSN&\textbf{1.94e-7}&\textbf{6.06e-7}&\textbf{8.10e-7}&\underline{\textbf{1.19e-7}}&\underline{\textbf{1.04e-7}}&5.13e-7\\
\hline
\multirow{3}{*}{375}&PINN&1.97e-6&2.52e-7&1.24e-6&1.08e-6&5.90e-7&1.02e-6\\
&PISN&3.18e-5&5.21e-6&3.05e-6&4.65e-6&1.15e-6&1.69e-6\\
&PINSN&\underline{\textbf{1.38e-7}}&\underline{\textbf{3.18e-8}}&\underline{\textbf{2.71e-8}}&\textbf{3.66e-7}&\textbf{6.84e-8}&\textbf{4.71e-8}\\
\hline
\multirow{3}{*}{475}&PINN&4.34e-6&3.65e-6&1.40e-6&7.72e-6&1.09e-7&4.88e-7\\
&PISN&1.79e-5&1.97e-6&1.75e-6&1.03e-6&3.72e-6&9.01e-6\\
&PINSN&\underline{\textbf{1.54e-8}}&\underline{\textbf{2.52e-8}}&\textbf{3.90e-8}&\textbf{2.25e-8}&\textbf{8.14e-8}&\underline{\textbf{8.66e-9}}\\
\hline
\multirow{3}{*}{725}&PINN&6.31e-7&9.20e-8&1.45e-7&7.22e-8&7.93e-8&3.01e-8\\
&PISN&2.27e-5&1.46e-6&1.56e-6&7.74e-5&3.92e-6&1.09e-6\\
&PINSN&\underline{\textbf{1.76e-7}}&\textbf{1.00e-8}&\textbf{6.93e-8}&\textbf{1.26e-8}&\underline{\textbf{3.36e-9}}&\underline{\textbf{8.41e-9}}\\
\hline
\multirow{3}{*}{975}&PINN&4.21e-6&1.97e-7&3.61e-6&2.53e-6&1.09e-7&4.86e-7\\
&PISN&1.08e-5&1.18e-6&7.80e-6&9.00e-6&3.22e-6&9.76e-6\\
&PINSN&\textbf{6.83e-7}&\underline{\textbf{1.02e-8}}&\textbf{4.22e-8}&\underline{\textbf{6.40e-8}}&\textbf{8.34e-8}&\underline{\textbf{2.29e-8}}\\
\hline
\end{tabular}
\vspace{1pt}
\label{tab:Navier-Stokes-Hypernetwork-Mean}
\end{table}

\begin{figure}
    \centering
    \includegraphics[width=\textwidth,height=\textheight]{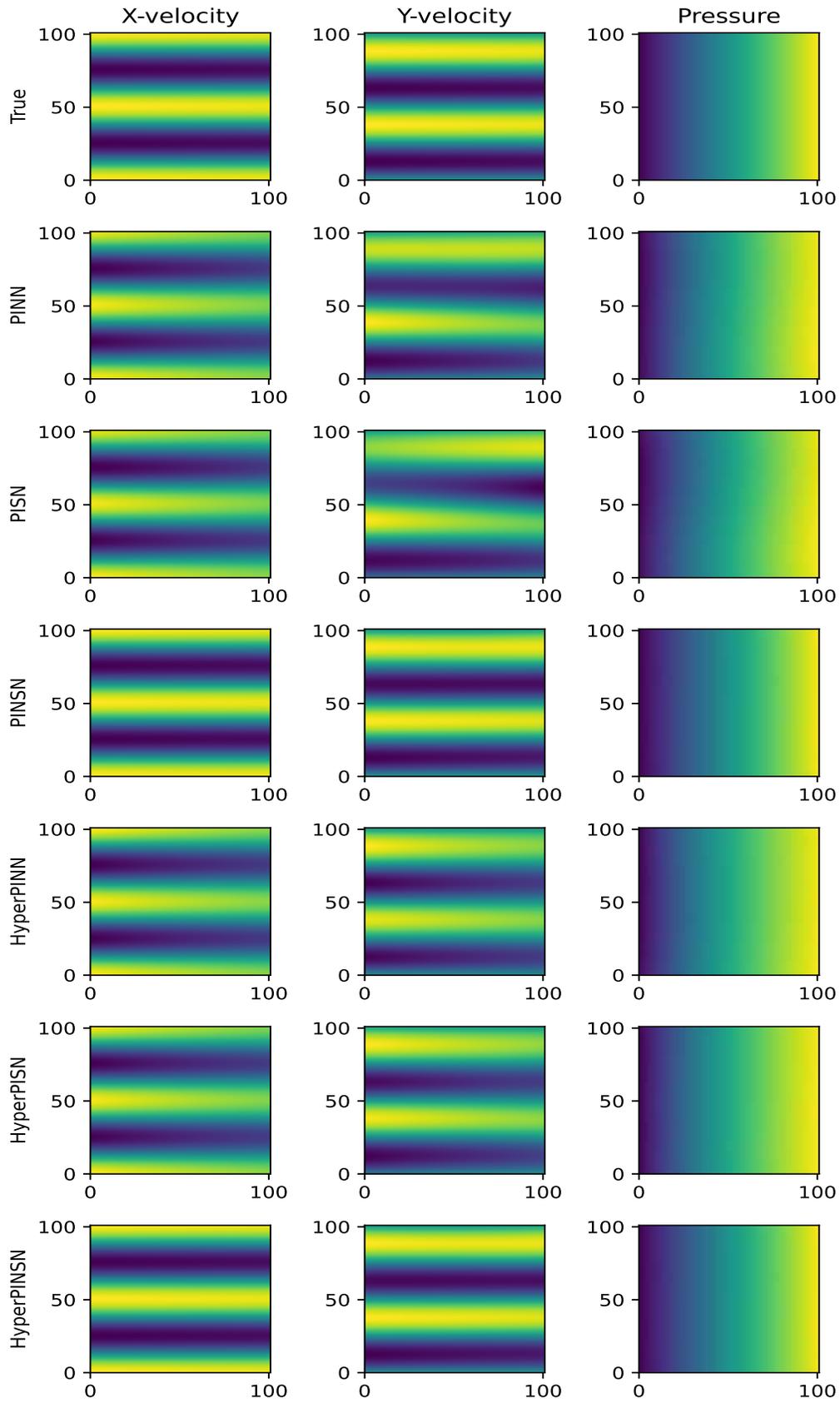}
    \caption{Navier-Stokes heat-maps for Reynolds Number=475}
    \label{fig:NS}
\end{figure}

\begin{figure}
    \centering
    \includegraphics[height=\textheight]{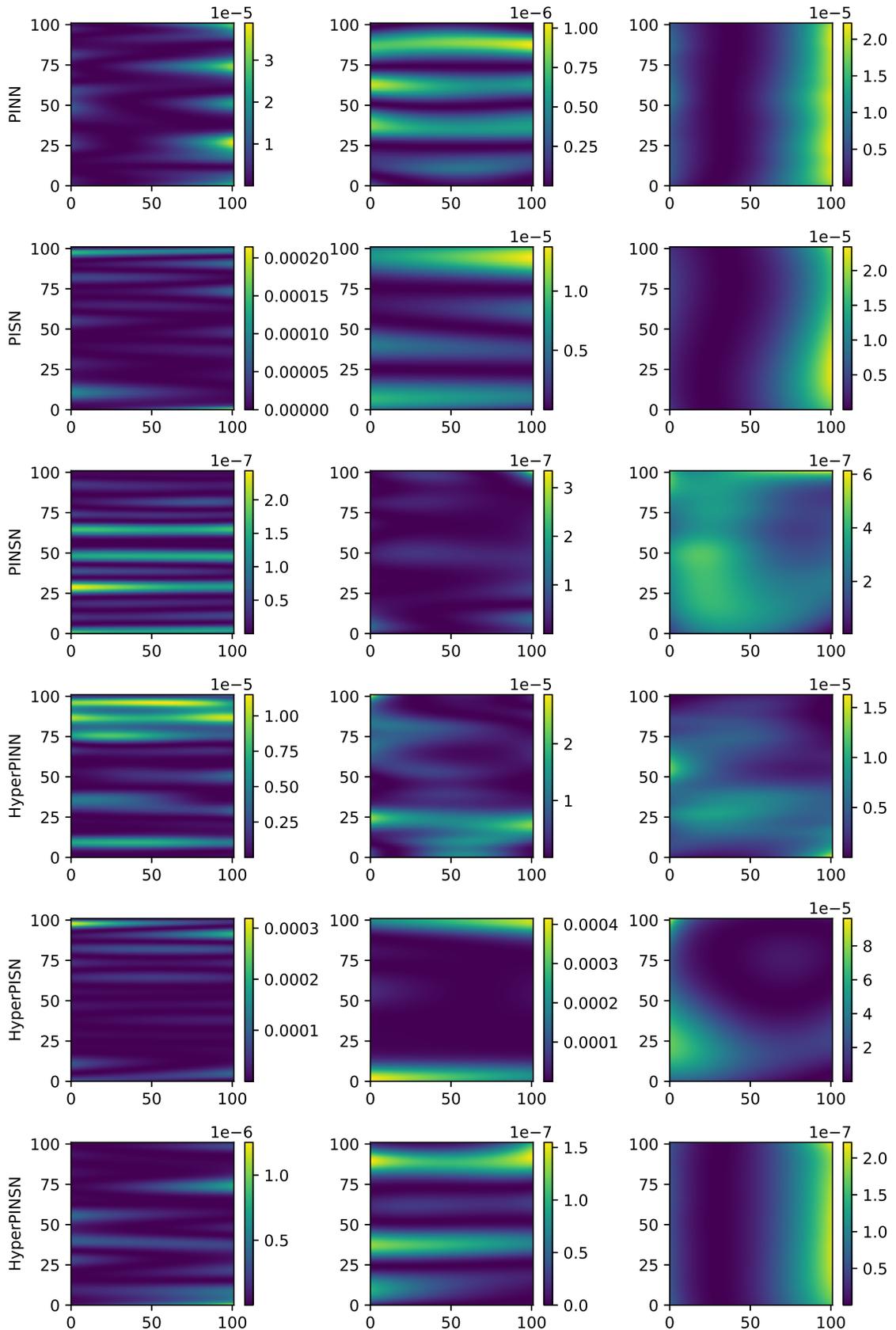}
    \caption{Error Maps for Navier-Stokes Reynolds Number=475}
    \label{fig:Error_NS}
\end{figure}

\begin{table}
    \caption{Kovasznay flow (Navier-Stokes): Hyperparameter details}
    \label{tab:NS-Hyperparameters}
    \centering
    \begin{tabular}{c c c c c c}
    &Optimizer&Starting-lr&$\gamma$&Decay schedule &Hidden-layers\\ 
    \hline
    PINN & Adam & $1e^{-4}$ & 0.1 & [40k,80k,120k]&[20*6]\\
    PISN & Adam & $1e^{-2}$ & 0.1 & [25k,50k,75k,100k]& Section \ref{Symbolic_Networks} \\
    PINSN & Adam & $1e^{-4}$ & 0.1 & [40k,80k,120k]&[20*6]\\
    Hypernetworks & Adam & $1e^{-4}$ & 0.1 & [50k,100k]&[512*2-256*2-128]\\
    \hline
\end{tabular}
\end{table}

\subsubsection{2D Coupled Burger's Equation}
\label{2D-Coupled-Burgers-details}

Table \ref{2D-Burger's:PDE Information} represents the governing differential equations and true analytical solutions of 2D-Coupled Burger's equation. The two differential equations represent momentum equations in X and Y directions. For modeling the temporal component, we consider a timestep interval of 0.1 for $t\in[0,1]$. The solution space is decomposed into 12 components by splitting the grid at $X = [0.4,0.8]$ and $Y=[0.25,0.50,0.75]$ , whose respective MLP architectures and residual points are mentioned in Table \ref{Domain-decomposition-hyperparameters}. Remaining training details are identical to that used for Kovasznay flow. Table \ref{tab:2D-Burger's-2-output-Domain_Hypernetwork-Mean} represents the mean-L2-error results for the tasks considered in Section \ref{section:Coupled-Burgers}. Tables \ref{tab:2D-Burger's-Domain-Mean} and \ref{tab:2D-Burger's-Domain-Max} represent domainwise mean-L2-error and max-L2-error results for $\nu=4.3e-3$ respectively. All experiments were conducted on Nvidia P100 GPU with 16 GB GPU Memory and 1.32 GHz GPU Memory clock, and Pytorch framework was used.

\begin{table}
  \caption{Two-dimensional coupled Burger's equation}
  \label{2D-Burger's:PDE Information}
  \centering
  \begin{tabular}{c c}
  \hline
  Governing Differential Equations & True Analytical solutions\\
  \hline
  \(u_{t} + u*u_{x} + v*u_{y} = \nu*(u_{xx} + u_{yy})\)&\(u(x,y,t) = 3/4+1/{4(1+exp[(-4x+4y-t)/32\nu])}\)\\
  \(v_{t} + u*v_{x} + v*v_{y} = \nu*(v_{xx} + v_{yy})\)&\(v(x,y,t) = 3/4-1/{4(1+exp[(-4x+4y-t)/32\nu])}\)\\
  \hline
  \end{tabular}
\end{table}

\begin{table}
    \centering
    \caption{Domain-decomposition hyperparameters for 2D-Burger's Equation}
    \begin{tabular}{c c c c c c}
    Domain Number&Range of X&Range of Y&\#Layers&\#Neurons&\#Residual Points\\
    \hline
    1&[0,0.4]&[0,0.25]&6&20&1000\\
    2&[0.4,0.8]&[0,0.25]&4&20&600\\
    3&[0.8,1.0]&[0,0.25]&2&20&400\\
    4&[0,0.4]&[0.25,0.5]&6&20&1000\\
    5&[0.4,0.8]&[0.25,0.5]&6&20&1000\\
    6&[0.8,1.0]&[0.25,0.5]&2&20&600\\
    7&[0,0.4]&[0.5,0.75]&4&20&400\\
    8&[0.4,0.8]&[0.5,0.75]&6&20&1000\\
    9&[0.8,1.0]&[0.5,0.75]&4&20&600\\
    10&[0,0.4]&[0.75,1.0]&2&20&400\\
    11&[0.4,0.8]&[0.75,1.0]&6&20&1000\\
    12&[0.8,1.0]&[0.75,1.0]&6&20&1000\\
    \hline
    \end{tabular}
    \label{Domain-decomposition-hyperparameters}
\end{table}

\begin{table}
\centering
\caption{2D Burger's: pointwise-mean-L2-Error viz-a-viz true solution}
\begin{tabular}{c c c c c c}
&&\multicolumn{2}{c}{Hypernetwork}&\multicolumn{2}{c}{Vanilla}\\
\hline
$\nu$ (1e-3)&Architecture&X-velocity&Y-velocity&X-velocity&Y-velocity\\
\hline
\multirow{3}{*}{2.2}&PINN&4.17e-5&9.22e-4&1.14e-5&5.44e-4\\
&PISN&8.53e-4&3.58e-3&2.59e-4&8.07e-3\\
&PINSN&\underline{\textbf{1.24e-5}}&\underline{\textbf{4.97e-5}}&\underline{\textbf{1.19e-5}}&\underline{\textbf{2.06e-5}}\\
\hline
\multirow{3}{*}{4.3}&PINN&1.33e-5&1.67e-4&1.48e-5&1.39e-4\\
&PISN&2.39e-4&1.08e-4&1.76e-4&1.77e-4\\
&PINSN&\underline{\textbf{3.57e-6}}&\underline{\textbf{1.94e-5}}&\underline{\textbf{6.51e-6}}&\underline{\textbf{3.45e-6}}\\
\hline
\multirow{3}{*}{5.8}&PINN&4.58e-5&7.72e-4&4.54e-5&7.52e-4\\
&PISN&7.79e-4&8.16e-3&7.78e-4&8.09e-3\\
&PINSN&\underline{\textbf{6.46e-6}}&\underline{\textbf{1.15e-4}}&\underline{\textbf{5.25e-6}}&\underline{\textbf{5.17e-5}}\\
\hline
\multirow{3}{*}{7.5}&PINN&8.81e-5&4.61e-4&8.76e-5&4.31e-4\\
&PISN&9.91e-4&1.98e-3&1.00e-3&1.88e-3\\
&PINSN&\underline{\textbf{1.14e-5}}&\underline{\textbf{2.19e-4}}&\underline{\textbf{1.33e-5}}&\underline{\textbf{2.42e-4}}\\
\hline
\multirow{3}{*}{9.3}&PINN&5.45e-4&1.15e-4&5.37e-4&7.50e-5\\
&PISN&1.92e-3&5.19e-3&1.91e-3&5.16e-3\\
&PINSN&\underline{\textbf{1.49e-5}}&\underline{\textbf{4.42e-5}}&\underline{\textbf{8.63e-5}}&\underline{\textbf{3.36e-5}}\\
\hline
\end{tabular}
\vspace{1pt}
\label{tab:2D-Burger's-2-output-Domain_Hypernetwork-Mean}
\end{table}

\begin{figure}
    \centering
    \includegraphics[width=\textwidth,height=\textheight]{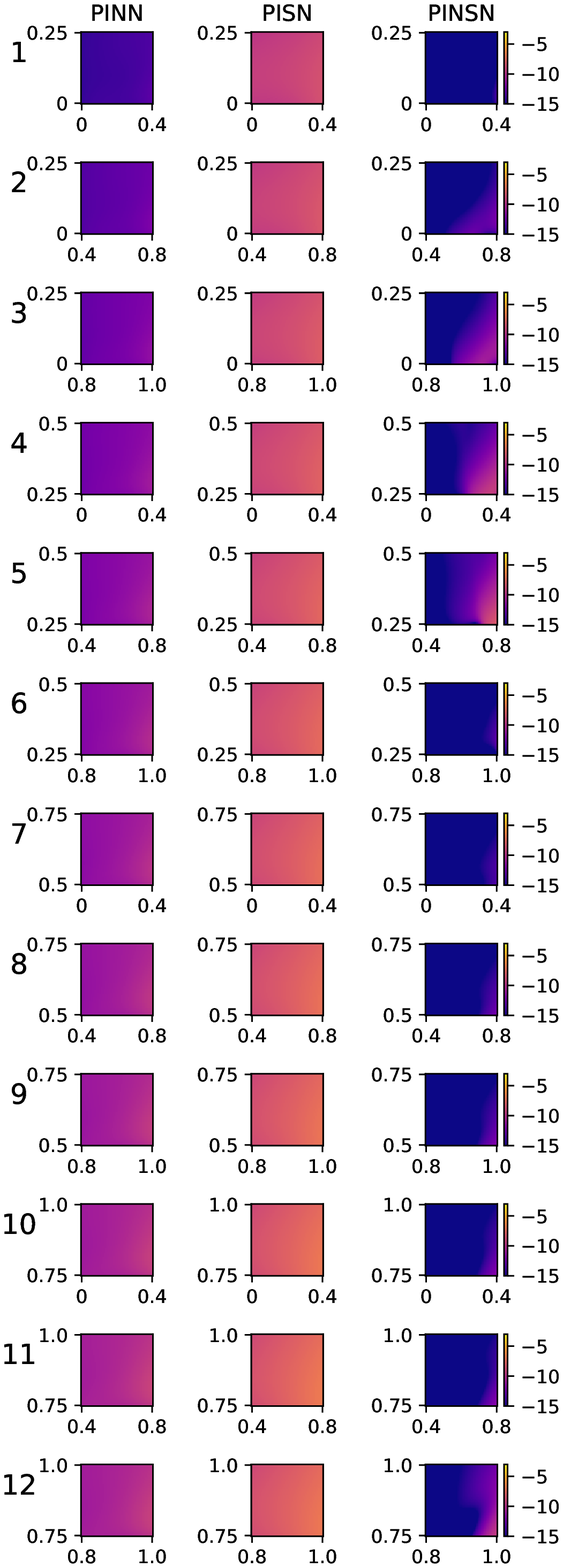}
    \caption{Domainwise error-maps in logarithmic scale for Vanilla architectures on X-velocity of 2D Burger's Equation with $\nu=4.3e-3$}
    \label{fig:Burger's-Vanilla-u}
\end{figure}

\begin{figure}
    \centering
    \includegraphics[width=\textwidth,height=\textheight]{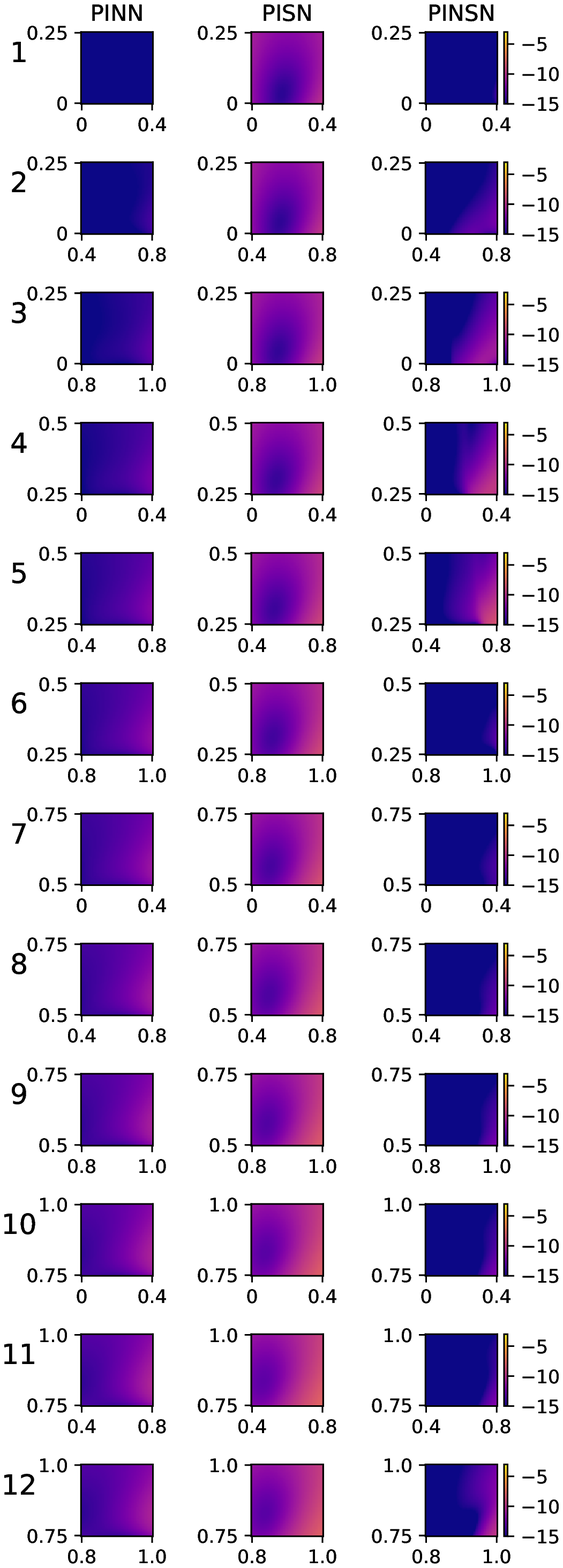}
    \caption{Domainwise error-maps in logarithmic scale for Vanilla architectures on Y-velocity of 2D Burger's Equation with $\nu=4.3e-3$}
    \label{fig:Burger's-Vanilla-v}
\end{figure}

\begin{figure}
    \centering
    \includegraphics[width=\textwidth,height=\textheight]{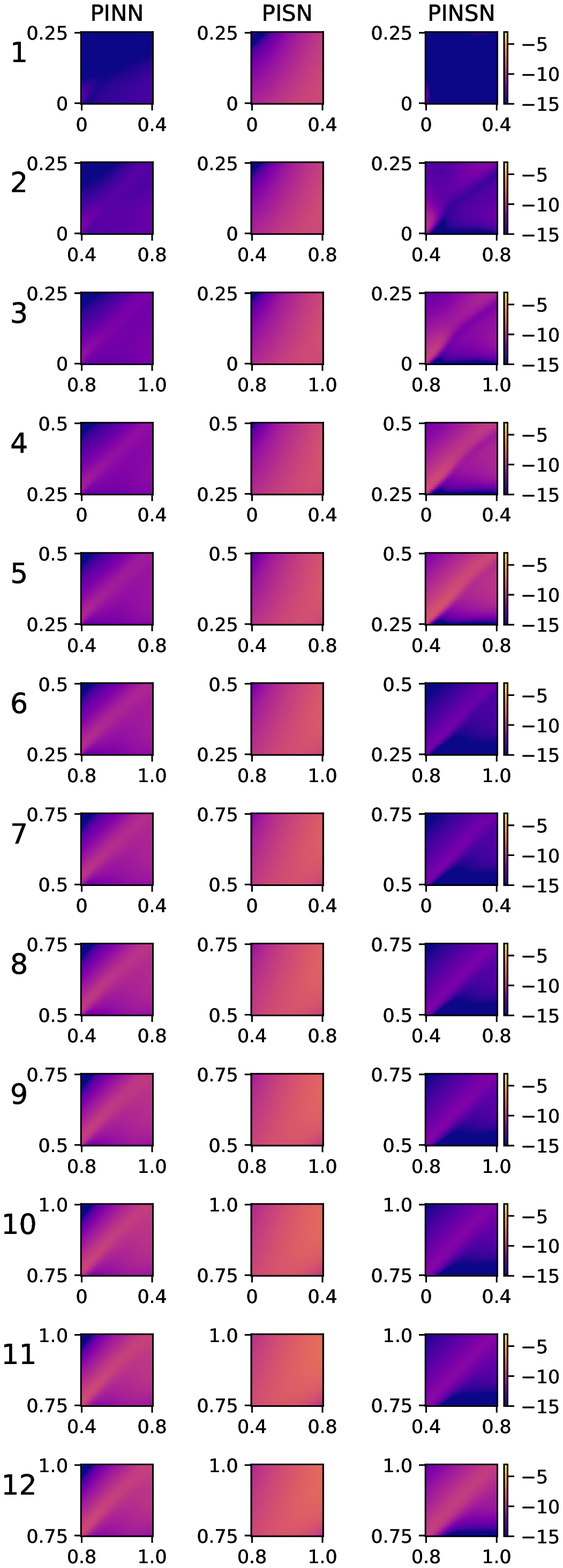}
    \caption{Domainwise error-maps in logarithmic scale for hyper architectures on X-velocity of 2D Burger's Equation with $\nu=4.3e-3$}
    \label{fig:Burger's-Hyper-u}
\end{figure}

\begin{figure}
    \centering
    \includegraphics[width=\textwidth,height=\textheight]{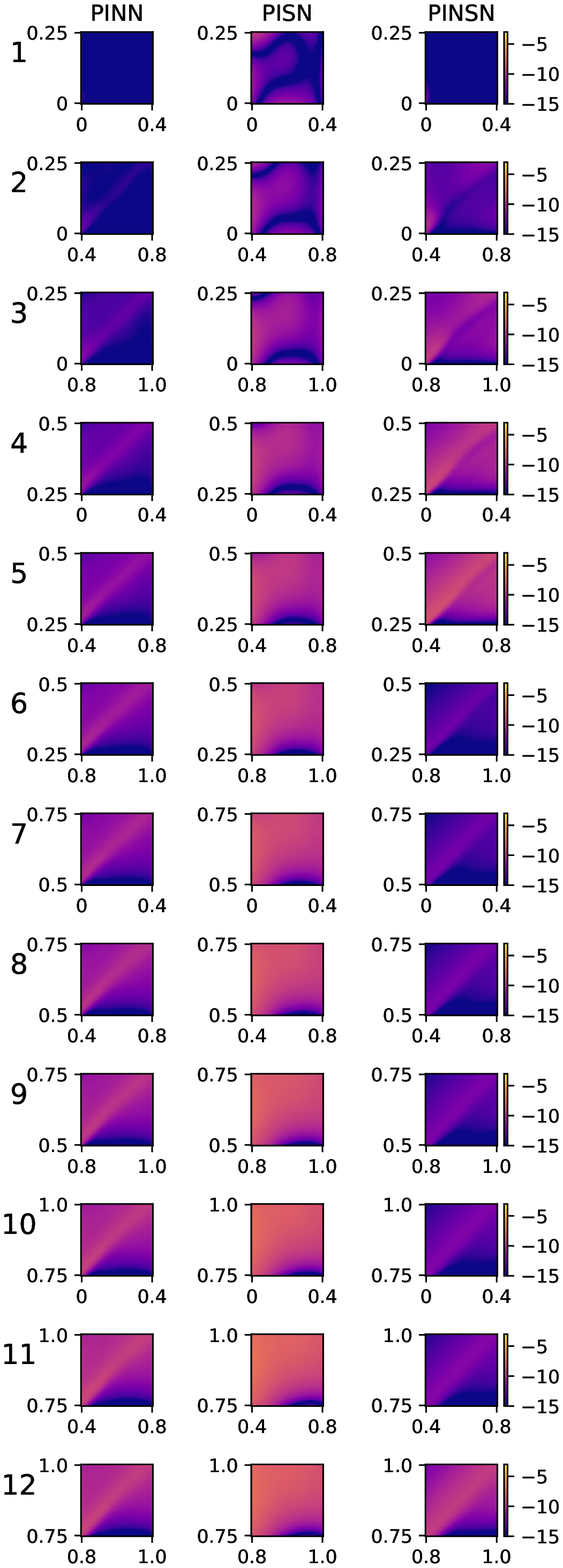}
    \caption{Domainwise error-maps in logarithmic scale for hyper architectures on Y-velocity of 2D Burger's Equation with $\nu=4.3e-3$}
    \label{fig:Burger's-Hyper-v}
\end{figure}

\begin{table}
\centering
\caption{Domainwise-Max-L2-Error for 2D Burger's at $\nu$ = 4.3e-3}
\begin{tabular}{c c c c c c c c c}
&&\multicolumn{3}{c}{X-velocity}&\multicolumn{3}{c}{Y-velocity}\\
\hline
&Domain&PISN&PINSN&PINN&PISN&PINSN&PINN\\
\hline
\multirow{12}{*}{Hypernetwork}&1&5.66e-4&\textbf{2.36e-5}&6.72e-4&1.62e-4&\textbf{4.14e-5}&2.26e-3\\
&2&2.62e-3&\underline{\textbf{1.69e-5}}&8.82e-4&2.03e-4&\underline{\textbf{4.12e-5}}&1.82e-4\\
&3&4.28e-4&\underline{\textbf{5.58e-5}}&5.56e-4&1.73e-4&\textbf{4.26e-5}&1.72e-4\\
&4&6.62e-6&7.12e-8&\underline{\textbf{7.02e-8}}&5.05e-7&\textbf{6.07e-8}&6.23e-6\\
&5&1.05e-5&\underline{\textbf{1.03e-6}}&2.32e-6&1.31e-5&\underline{\textbf{3.12e-6}}&1.50e-4\\
&6&1.26e-3&\textbf{6.76e-5}&7.19e-5&7.74e-4&\underline{\textbf{1.64e-5}}&6.96e-5\\
&7&7.26e-3&\underline{\textbf{3.11e-5}}&5.71e-4&3.29e-4&\underline{\textbf{1.08e-5}}&2.96e-4\\
&8&1.98e-3&\underline{\textbf{1.12e-5}}&7.21e-4&1.11e-4&\underline{\textbf{1.86e-5}}&1.32e-4\\
&9&1.33e-2&\textbf{1.71e-4}&5.51e-4&1.25e-5&\underline{\textbf{3.25e-4}}&8.77e-5\\
&10&5.03e-4&\textbf{3.38e-5}&3.16e-4&2.72e-5&\underline{\textbf{5.19e-6}}&2.42e-5\\
&11&1.37e-3&\textbf{7.47e-5}&7.72e-5&4.11e-5&\underline{\textbf{1.28e-6}}&2.82e-5\\
&12&8.08e-3&\underline{\textbf{4.42e-5}}&3.91e-4&1.19e-2&\textbf{1.95e-4}&2.18e-2\\
\hline
\multirow{12}{*}{Vanilla}&1&2.07e-4&\underline{\textbf{2.03e-5}}&6.61e-4&1.30e-4&\underline{\textbf{3.82e-5}}&2.63e-3\\
&2&5.11e-3&\textbf{4.71e-5}&7.37e-4&2.15e-4&\textbf{5.54e-5}&5.97e-5\\
&3&4.87e-4&\textbf{7.02e-5}&4.98e-4&1.86e-4&\underline{\textbf{1.36e-5}}&9.61e-5\\
&4&3.17e-6&\textbf{8.72e-7}&7.52e-6&5.28e-7&\underline{\textbf{4.65e-8}}&1.44e-7\\
&5&2.43e-5&\textbf{4.19e-6}&8.36e-6&3.84e-5&\textbf{6.31e-6}&3.88e-5\\
&6&5.56e-3&\underline{\textbf{5.44e-5}}&7.47e-5&5.93e-4&\textbf{7.86e-5}&4.52e-5\\
&7&6.09e-3&\textbf{4.42e-5}&5.52e-4&3.55e-4&\textbf{4.91e-5}&3.77e-4\\
&8&4.49e-3&\textbf{2.73e-5}&3.89e-4&1.96e-4&\textbf{2.06e-5}&1.56e-4\\
&9&3.97e-2&\underline{\textbf{9.16e-5}}&5.85e-4&1.93e-2&\textbf{1.66e-3}&4.37e-2\\
&10&5.13e-3&\underline{\textbf{1.16e-5}}&3.26e-5&3.71e-5&\textbf{5.57e-6}&2.81e-5\\
&11&6.81e-4&7.63e-5&\underline{\textbf{5.17e-5}}&4.24e-5&\textbf{4.86e-6}&3.61e-5\\
&12&2.26e-4&\textbf{4.82e-5}&3.86e-4&1.59e-2&\underline{\textbf{1.71e-4}}&2.58e-3\\
\hline
\end{tabular}
\vspace{1pt}
\label{tab:2D-Burger's-Domain-Max}
\end{table}

\begin{table}
\centering
\caption{Domain-wise Mean-L2-Error for 2D Burger's at $\nu$ = 4.3e-3}
\begin{tabular}{c c c c c c c c}
&&\multicolumn{3}{c}{X-velocity}&\multicolumn{3}{c}{Y-velocity}\\
\hline
&Domain&PISN&PINSN&PINN&PISN&PINSN&PINN\\
\hline
\multirow{12}{*}{Hypernetwork}&1&7.07e-5&3.46e-6&\textbf{3.06e-6}&2.02e-4&\textbf{5.19e-6}&2.82e-4\\
&2&1.86e-4&\underline{\textbf{1.37e-6}}&1.22e-5&1.45e-3&\underline{\textbf{3.04e-5}}&1.31e-4\\
&3&3.89e-5&\underline{\textbf{1.64e-6}}&1.65e-6&1.17e-5&\textbf{3.96e-6}&1.57e-5\\
&4&5.44e-7&\underline{\textbf{2.21e-8}}&3.56e-8&4.12e-8&\underline{\textbf{1.16e-8}}&5.91e-7\\
&5&1.89e-6&\underline{\textbf{1.11e-7}}&1.12e-7&1.25e-6&\textbf{2.54e-7}&1.52e-5\\
&6&1.57e-4&9.02e-6&\textbf{8.42e-6}&9.67e-5&\underline{\textbf{1.67e-5}}&8.70e-4\\
&7&5.58e-4&\textbf{2.26e-6}&2.82e-5&2.53e-4&\textbf{8.49e-5}&2.27e-4\\
&8&1.65e-4&9.16e-6&\textbf{9.11e-6}&9.21e-5&\textbf{1.52e-5}&1.11e-4\\
&9&1.10e-3&\textbf{4.55e-6}&5.84e-5&1.03e-4&\underline{\textbf{2.64e-6}}&7.21e-6\\
&10&6.29e-5&\underline{\textbf{3.77e-6}}&4.04e-6&3.36e-6&\textbf{6.76e-7}&3.02e-5\\
&11&1.05e-4&5.89e-6&\underline{\textbf{5.80e-6}}&3.14e-6&\underline{\textbf{6.36e-8}}&2.19e-7\\
&12&5.77e-4&\textbf{2.85e-6}&2.98e-5&2.84e-3&\underline{\textbf{7.27e-5}}&3.12e-4\\
\hline
\multirow{12}{*}{Vanilla}&1&2.27e-4&\underline{\textbf{1.65e-6}}&1.91e-6&2.18e-4&\underline{\textbf{3.74e-6}}&1.99e-4\\
&2&1.89e-4&\textbf{1.55e-6}&5.16e-5&1.07e-3&\textbf{3.86e-5}&4.49e-4\\
&3&3.94e-5&\textbf{2.93e-6}&4.41e-6&1.52e-5&\underline{\textbf{2.91e-6}}&8.84e-6\\
&4&8.15e-7&\textbf{2.52e-8}&3.51e-8&7.75e-8&\textbf{1.73e-8}&3.91e-7\\
&5&5.21e-6&\textbf{3.56e-7}&6.82e-7&1.24e-6&\underline{\textbf{2.22e-7}}&1.19e-5\\
&6&1.08e-4&7.18e-6&\underline{\textbf{5.32e-6}}&5.52e-5&\textbf{1.92e-5}&4.74e-5\\
&7&2.39e-4&\underline{\textbf{1.85e-6}}&4.59e-5&3.77e-4&\underline{\textbf{6.75e-5}}&1.97e-4\\
&8&1.12e-4&7.45e-6&\underline{\textbf{5.12e-6}}&5.19e-5&\underline{\textbf{1.37e-5}}&1.39e-4\\
&9&4.18e-4&\underline{\textbf{2.75e-6}}&1.87e-5&1.36e-4&\textbf{3.64e-6}&5.12e-6\\
&10&8.94e-5&\textbf{6.12e-6}&9.73e-6&1.04e-5&\underline{\textbf{4.43e-7}}&3.55e-5\\
&11&3.29e-4&9.15e-6&\textbf{7.08e-6}&5.24e-6&\underline{\textbf{6.97e-8}}&2.95e-7\\
&12&3.62e-4&\underline{\textbf{2.37e-6}}&2.84e-5&1.67e-4&\textbf{7.46e-5}&5.76e-4\\
\hline
\end{tabular}
\vspace{1pt}
\label{tab:2D-Burger's-Domain-Mean}
\end{table}

\subsection{Additional Differential Equations}
\subsubsection{Telegraph Equations}

\begin{table}
    \centering
    \caption{Telegraph Equations: PDE Information}
    \begin{tabular}{ c c c c c}
          & f(x,0) & g(x,0) & Range & True Solution \\
         \hline
         \rule{0pt}{2ex} Telegraph-1:
         &\multirow{2}{*}{\(Ae^x\)} &\multirow{2}{*}{\(-(A+1)e^x\)}
         & \multirow{2}{*}{[0.5,5]}
         & \multirow{2}{*}{\(e^{Ax-(A+1)t}\)}\\
         \(u_{xx}= u_{tt}+2u_t+u\)&&&&\\
         \hline
         \rule{0pt}{2ex} Telegraph-2:
         &\multirow{2}{*}{\(1+e^{Ax}\)} 
         &\multirow{2}{*}{\(-A\)}
         & \multirow{2}{*}{[0.2,3.2]}
         & \multirow{2}{*}{\(e^{Ax}+e^{-At}\)}\\
         \(u_{xx}=u_{tt}+2Au_t+A^2u\)& &&&\\
         \hline
    \end{tabular}
    \vspace{2pt}
    \label{tab:Telegraph_information}
\end{table}

Telegraph equations \cite{Telegraph} are a sub-derivative of Maxwell equations, used to capture the variations in current and voltage which vary both spatially and temporally. Telegraph equations are extensively used in modeling transmission lines of varying frequencies, radio frequency conductors, telephone lines, and pulses of direct current. Table  \ref{tab:Telegraph_information} contains the information for parameterized PDEs of Telegraph equation. Table \ref{tab:Telegraph_information} represents the PDE information of Telegraph Equations. Initial Dirichlet and Neumann condition is represented using f(x) and g(x) respectively, with u(x,0)=f(x) and \(\frac{\partial u(x,0)}{\partial t}\) = g(x). The variable $A\in R^1$  is used to represent the parameterization of the initial conditions, and PDE (in Telegraph-2) and the range column defines the range of values A can take. The final column states the true analytical solution of each of the PDE problems, parameterized by A.

\textbf{Training Schedule and Hyperparameters:} The Telegraph equation is 1-Dimensional $x\in[0,1]$ and varying in time $t\in[0,1]$. We consider 2601 points as train-collocation points, and sample 40 initial and boundary condition points respectively. Remaining training schedule details are identical to previous experiments. Tables \ref{tab:Telegraph-Mean} and \ref{tab:Telegraph-Max} represents the mean-L2-error and max-L2-error results for Telegraph Equations. We observe, (Hyper)PINSNs boost the accuracy by 2-3 orders of magnitude than (Hyper)PISNs and (Hyper)PINNs.

\begin{table}
\centering
\caption{Telegraph Equations: Pointwise-Mean-L2-Error Results}
\begin{tabular}{c c c c c c c c}
&&\multicolumn{3}{c}{Hypernetwork}&\multicolumn{3}{c}{Vanilla}\\
\hline
&Task&PINN&PISN&PINSN&PINN&PISN&PINSN\\
\hline
\multirow{5}{*}{Telegraph-1}&0.7&5.22e-5&8.23e-4&\textbf{1.95e-6}&4.46e-5&7.87e-4&\underline{\textbf{1.78e-6}}\\
&1.5&2.31e-5&1.39e-4&\textbf{2.66e-6}&2.02e-5&1.38e-4&\underline{\textbf{2.45e-6}}\\
&2.4&7.06e-5&1.08e-4&\textbf{6.17e-6}&6.16e-5&1.05e-4&\underline{\textbf{5.62e-6}}\\
&3.6&1.58e-5&7.76e-4&\textbf{3.22e-6}&1.33e-5&7.37e-4&\underline{\textbf{2.93e-6}}\\
&4.7&5.07e-5&5.19e-4&\textbf{3.59e-6}&4.40e-5&5.03e-4&\underline{\textbf{3.36e-6}}\\
\hline
\multirow{5}{*}{Telegraph-2}&0.27&4.74e-4&3.70e-4&\textbf{3.75e-5}&3.88e-4&2.63e-4&\underline{\textbf{3.24e-5}}\\
&0.51&6.09e-4&7.28e-4&\textbf{4.51e-5}&8.91e-4&3.47e-4&\underline{\textbf{3.59e-5}}\\
&0.88&2.32e-4&6.24e-4&\textbf{2.96e-5}&1.99e-4&2.64e-4&\underline{\textbf{2.66e-5}}\\
&1.25&4.66e-4&4.77e-4&\textbf{5.85e-6}&3.83e-4&4.39e-4&\underline{\textbf{5.42e-6}}\\
&1.71&1.10e-4&6.47e-4&\textbf{3.38e-5}&9.00e-4&3.79e-4&\underline{\textbf{2.75e-5}}\\
\hline
\end{tabular}
\vspace{1pt}
\label{tab:Telegraph-Mean}
\end{table}

\begin{table}
\centering
\caption{Telegraph Equations Pointwise-Max-L2-Error Results}
\begin{tabular}{c c c c c c c c}
&&\multicolumn{3}{c}{Hypernetwork}&\multicolumn{3}{c}{Vanilla}\\
\hline
&Task&PINN&PISN&PINSN&PINN&PISN&NS-PINN\\
\hline
\multirow{5}{*}{Telegraph-1}&0.7&6.12e-4&1.15e-3&\textbf{3.79e-6}&6.69e-4&8.66e-3&\underline{\textbf{3.64e-6}}\\
&1.5&1.85e-4&2.09e-3&\underline{\textbf{5.30e-6}}&2.02e-4&1.79e-3&\textbf{6.41e-5}\\
&2.4&7.77e-4&1.62e-3&\textbf{3.19e-5}&5.54e-4&1.47e-3&\underline{\textbf{1.29e-5}}\\
&3.6&1.90e-4&1.01e-3&\underline{\textbf{6.86e-6}}&1.86e-4&9.58e-3&\textbf{1.04e-5}\\
&4.7&5.58e-4&5.71e-3&\textbf{7.57e-5}&4.84e-4&7.55e-3&\underline{\textbf{1.78e-5}}\\
\hline
\multirow{5}{*}{Telegraph-2}&0.27&6.16e-3&3.70e-3&\textbf{1.68e-4}&3.49e-3&2.37e-3&\underline{\textbf{9.49e-5}}\\
&0.51&8.53e-3&8.74e-3&\textbf{1.81e-4}&8.02e-3&5.21e-3&\underline{\textbf{1.51e-4}}\\
&0.88&3.48e-3&5.62e-3&\textbf{1.63e-4}&2.79e-3&2.38e-3&\underline{\textbf{1.48e-4}}\\
&1.25&5.59e-3&5.25e-3&\textbf{2.01e-5}&3.83e-3&3.95e-3&\underline{\textbf{1.56e-5}}\\
&1.71&1.43e-3&9.71e-3&\underline{\textbf{1.71e-4}}&9.90e-3&5.31e-3&\textbf{1.93e-4}\\
\hline
\end{tabular}
\vspace{1pt}
\label{tab:Telegraph-Max}
\end{table}

\subsubsection{2D Burger's: Conservation Equation}

The two-dimensional viscous Burger's equation in it's conservation form is given by: \[ u_{t} + p_{x} + q_{y} = \nu(u_{xx} + u_{yy}) \] Where \(p(u)=u^2/2\) and \(q(u)=u^2/2\) are the inviscid fluxes, and $\nu$ represents the Viscosity coefficient governing the diffusion component of the PDE. It's analytical solution is given by \(u(x,y,t) = (1+exp(x+y-t)/2\nu)^{-1}\).

\begin{table}
\centering
\caption{Domain Decomposed HyperNetwork Results: 2D Burger's: Conservation Equation}
\begin{tabular}{c c c c c c}
&&\multicolumn{2}{c}{Hypernetwork}&\multicolumn{2}{c}{Vanilla}\\
\hline
$\nu$ &Architecture&u-Mean&u-Max&u-Mean&u-Max\\
\hline
\multirow{3}{*}{5.30e-3}&PINN&3.71e-4&1.52e-3&3.62e-4&1.52e-3\\
&PISN&5.48e-4&4.71e-3&5.51e-4&4.71e-3\\
&PINSN&\textbf{3.25e-6}&\textbf{3.19e-5}&\underline{\textbf{1.89e-6}}&\underline{\textbf{2.67e-5}}\\
\hline
\multirow{3}{*}{9.40e-3}&PINN&3.15e-5&1.15e-3&2.65e-5&1.15e-3\\
&PISN&1.15e-4&9.81e-2&1.19e-4&9.81e-2\\
&PINSN&\textbf{5.09e-5}&\textbf{3.19e-4}&\underline{\textbf{5.68e-5}}&\underline{\textbf{3.77e-4}}\\
\hline
\multirow{3}{*}{1.10e-2}&PINN&2.24e-4&1.59e-3&2.16e-4&1.58e-3\\
&PISN&2.29e-4&6.51e-3&2.27e-4&6.50e-3\\
&PINSN&\textbf{3.81e-5}&\textbf{2.16e-4}&\underline{\textbf{4.64e-5}}&\underline{\textbf{3.81e-4}}\\
\hline
\multirow{3}{*}{2.30e-2}&PINN&8.22e-6&6.65e-4&7.22e-6&6.60e-4\\
&PISN&5.71e-5&9.22e-4&4.71e-5&9.20e-4\\
&PINSN&\textbf{2.02e-6}&\textbf{1.07e-5}&\underline{\textbf{2.96e-6}}&\underline{\textbf{1.18e-5}}\\
\hline
\multirow{3}{*}{3.70e-2}&PINN&5.51e-5&3.45e-4&4.81e-5&3.38e-4\\
&PISN&4.38e-4&2.29e-3&4.34e-4&2.28e-3\\
&PINSN&\textbf{8.85e-6}&\textbf{1.76e-5}&\underline{\textbf{4.24e-6}}&\underline{\textbf{1.65e-5}}\\
\hline
\end{tabular}
\vspace{1pt}
\label{tab:2D-Burger's-1-output-Domain_Hypernetwork}
\end{table}

The training schedule and hyperparameter specifications are identical to 2D-Coupled-Burger's experiment. The viscosity coefficient $\nu$ ranges between $[5e-3, 5e-2]$. Table \ref{tab:2D-Burger's-1-output-Domain_Hypernetwork} represents the results for domain-decomposed-hypernetwork and domain-decomposed-vanilla architectures. We observe that (Hyper) domain-decomposed-PINSNs outperform (Hyper) domain-decomposed-PINNs and (Hyper) domain-decomposed-PISNs by 1-2 orders of magnitude.

\subsection{Mathematical expressions generated in Appendix \ref{SymRegComparison}}
\label{Expressions}
\begin{table}
    \caption{Expressions generated by AI-Feynman and Symbolic-GPT for PDEs in Appendix \ref{SymRegComparison}}
    \label{expressions}
    \centering
    \begin{tabular}{c c}
    \hline
    AI-Feynman: Noiseless sample points& AI-Feynman: PINN-induced sample points\\
    \hline
    \vspace{1pt}
    \(sin(x)sin(t)\)&\(0.9988sin(x)sin(t)\)\\
    \(sin(x)exp(-t)\)&\(1.0002*sin(x)exp(-t)\)\\
    \(x+t\)&\(0.9999x+1.0001t\)\\
    \(xexp(t)\)&\(1.0001xexp(t)\)\\
    \((x+1)exp(t)\)&\(0.9998(x+1.00001)exp(0.9999t)\)\\
    \(0.94tan(0.38cos(3.13sin(3.14x))-0.47\)&\(0.95tan(0.52cos(2.98sin(3.14x))-0.66)\)\\
    \(1\)&\(4.42exp(x)\)\\
    \(0.5-0.5exp(\pi x)^{-0.25}\)&\(0.48-0.51exp(3.1312x)^{-0.25}\)\\
\(1.449exp(-0.91x+2.23t)\)&\(1.594exp(-0.95x+log(3.134x)1.478t)\) \\
\(3.774cos(0.52\pi x)\)&\(2.955sin(1.994\pi x)\)\\
    \hline
    \vspace{2pt}
    Symbolic-GPT: Noiseless sample points& Symbolic-GPT: PINN-induced sample points\\
    \hline
    \vspace{2pt}
\(-0.0006log(2.449t)^5+1.708\)&\(-0.0072log(3.884t)^4+2.248\)\\
\(-0.0029log(1.954t)^3+1.709\)&\(0.0052log(15.31t)^2+4.241\)\\
\(0.143x/(1.712x+0.320)+1.796\)&\(0.174x/(5.56x+0.6415)\)\\
\(-0.0029log(1.9547t)^3+1.7095\)&\(-0.0115log(2.041t)^7-3.317\)\\
\(-0.0006log(2.449t)^5+1.7085\)&\(-0.0006log(2.449t)^5+1.7085\)\\
\(0.007y-0.075log(2016.68y+926.36)+1.827\)&\(-0.124log(125.41y+2.26)+1.54\)\\
\(0.139x-0.0005y-0.011log(1.27y)+1.757\)&\(6.65x+4.41\)\\
\(0.1399x-0.0112log(1.3568y)+1.757\)&\(0.1725x-0.1256log(4.714y)-6.1042\)\\
\(-0.0029log(1.9547y)^3+1.709\)&\(-0.0048log(5.174y)^5\)\\
\(0.1399x+0.0004log(0.773y)^4+1.749\)&\(0.1581x+0.0071log(2.544y)^4+3.314\)\\
\hline
\end{tabular}
\end{table}

Mathematical Expression generated by HyperPISN for Kovasznay flow with Re=475:

X-velocity is given by:
\begin{equation}
    \begin{cases}
    l_1 = 0.006x-2.985y-0.066\\
    l_2 = -0.061x+0.845y+0.399\\
    l_3 = 0.017x-0.876y-0.614\\
    l_4 = -0.003x-0.713y-0.501\\
    l_5 = -0.122x+0.055y-2.458\\
    l_6 = -0.003x+1.258y-0.174\\
    h_{11} = 0.022x-0.685y+0.047sin(l_1)-0.22exp(l_2)-0.182(l_3+l_4)+1.366l_5l_6+0.634\\
    h_{12} = 0.217x-0.166y-1.319sin(l_1)-0.496exp(l_2)+0.489(l_3+l_4)+0.019l_5l_6+0.051\\
    h_{13} = 0.064x+0.144y-0.815sin(l_1)+0.194exp(l_2)-0.704(l_3+l_4)+0.721l_5l_6-0.011\\
    h_{14} = -0.532x-0.325y-0.506sin(l_1)-0.507exp(l_2)-0.653(l_3+l_4)+0.214l_5l_6+0.134\\
    h_{15} = -0.19x+1.113y-2.744sin(l_1)-0.401exp(l_2)-0.681(l_3+l_4)+0.467l_5l_6-0.913\\
    h_{16} = 0.001x+0.462y-1.720sin(l_1)-0.418exp(l_2)-1.157(l_3+l_4)+0.376l_5l_6-0.084\\
    u = -0.063x-0.115y+1.113sin(h_{11})+0.5exp(h_{12})-0.365(h_{13}+h_{14})+0.67h_{15}h_{16}-0.331\\
    \end{cases}
\end{equation}

Y-velocity is given by:
\begin{equation}
    \begin{cases}
    l_1 = 0.027x-0.078y-0.008\\
    l_2 = -0.035x-0.076y+0.488\\
    l_3 = 0.389x-0.407y-0.592\\
    l_4 = -0.18x+0.267y+0.225\\
    l_5 = -0.048x+2.071y+0.218\\
    l_6 = -0.04x-1.114y+0.639\\
    h_{11} = 0.017x-0.073y+0.208sin(l_1)-0.088exp(l_2)-0.198(l_3+l_4)+0.872l_5l_6-0.079\\
    h_{12} = 0.192x-0.215y+0.158sin(l_1)-0.332exp(l_2)-0.21(l_3+l_4)-0.757l_5l_6+0.067\\
    h_{13} =-0.036x-0.001y+0.295sin(l_1)+0.141exp(l_2)+0.098(l_3+l_4)-0.009l_5l_6+0.126\\
    h_{14} = -0.122x-0.11y-0.043sin(l_1)-0.19exp(l_2)+0.109(l_3+l_4)+0.137l_5l_6+0.063\\
    h_{15} = -0.075x+0.257y+0.128sin(l_1)+0.197exp(l_2)+0.018(l_3+l_4)+0.769l_5l_6+0.226\\
    h_{16} = 0.148x+0.167y-0.138sin(l_1)+0.106exp(l_2)-0.319(l_3+l_4)+0.764l_5l_6+0.062\\
    v = 3e^{-3}x+0.102y+0.24sin(h_{11})+0.16exp(h_{12})+3e^{-4}(h_{13}+h_{14})-0.23h_{15}h_{16}-0.024\\
    \end{cases}
\end{equation}

Pressure is given by:
\begin{equation}
    \begin{cases}
    l_1 = -0.177x+0.766y+1.469\\
    l_2 = -0.037x+0.018y+0.588\\
    l_3 = 0.03x+0.296y+0.623\\
    l_4 = 0.458x-0.142y-0.492\\
    l_5 = 0.115x-0.249y+0.623\\
    l_6 = 0.726x-0.402y+0.043\\
    h_{11} = -0.227x-0.328y+0.317sin(l_1)-0.301exp(l_2)+0.176(l_3+l_4)+0.341l_5l_6-0.192\\
    h_{12} = -0.249x-0.272y-0.403sin(l_1)+0.041exp(l_2)-0.308(l_3+l_4)+0.122l_5l_6-0.308\\
    h_{13} = -0.3x+0.115y+0.079sin(l_1)+0.45exp(l_2)+0.229(l_3+l_4)+0.464l_5l_6-0.151\\
    h_{14} = -0.267x+0.127y+0.152sin(l_1)+0.267exp(l_2)-0.308(l_3+l_4)+0.384l_5l_6-0.185\\
    h_{15} = 0.243x-0.327y-0.199sin(l_1)+0.325exp(l_2)+0.222(l_3+l_4)-0.088l_5l_6+0.215\\
    h_{16} = -0.007x-0.477y+0.257sin(l_1)+0.347exp(l_2)-0.09(l_3+l_4)-0.213l_5l_6-0.005\\
    p = 0.017x-0.045y-0.22sin(h_{11})-0.22exp(h_{12})+0.102(h_{13}+h_{14})+0.163h_{15}h_{16}-0.181\\
    \end{cases}
\end{equation}

\subsection{Impact of L-BFGS Optimizer and increasing the depth of Symbolic Network}

We used Adam Optimizer for our experiments in the main section. L-BFGS is a low memory version of Broyden–Fletcher–Goldfarb–Shanno algorithm, which is popularly used in PINN-based problems due to better convergence guarantees. In this section, we investigate the effects of applying L-BFGS optimizer on top of Adam Optimizer on one example each of Kovasznay flow and two-dimensional coupled viscous Burger's equation.

\begin{table}[h]
\centering
\caption{Pointwise-Max-L2-error on Kovasznay flow (Navier-Stokes) with Re=125}
\begin{tabular}{c c c c}
\hline
Model&X-velocity&Y-velocity&Pressure\\
\hline
PINN&7.74e-7&1.17e-7&7.48e-8\\
\hline
PISN-Depth-2&5.51e-5&7.72e-6&3.46e-6\\
PISN-Depth-3&2.94e-6&3.53e-7&3.04e-7\\
PISN-Depth-4&\textbf{1.54e-6}&\textbf{3.48e-7}&\textbf{3.01e-7}\\
\hline
PINSN-Depth-2&6.38e-7&\textbf{1.08e-8}&5.34e-8\\
PINSN-Depth-3&5.51e-7&7.96e-8&4.45e-8\\
PINSN-Depth-4&\textbf{5.29e-7}&{9.48e-7}&\textbf{3.95e-8}\\
\hline
\end{tabular}
\vspace{1pt}
\label{tab:L-BFGS-NS}
\end{table}

\begin{table}[h]
\centering
\caption{Pointwise-Max-L2-error on two-dimensional Burger's equation with $\nu$=2.2e-3}
\begin{tabular}{c c c c}
\hline
Model&X-velocity&Y-velocity\\
\hline
PINN&5.87e-5&1.27e-5\\
\hline
PISN-Depth-2&1.07e-3&5.59e-4\\
PISN-Depth-3&4.54e-4&3.53e-4\\
PISN-Depth-4&\textbf{4.38e-4}&\textbf{3.48e-4}\\
\hline
PINSN-Depth-2&{5.34e-5}&{1.19e-5}\\
PINSN-Depth-3&{8.46e-6}&{3.95e-6}\\
PINSN-Depth-4&\textbf{8.29e-6}&\textbf{4.42e-6}\\
\hline
\end{tabular}
\vspace{1pt}
\label{tab:L-BFGS-Burgers}
\end{table}

From Tables \ref{tab:L-BFGS-NS} and \ref{tab:L-BFGS-Burgers}, we observe, applying L-BFGS on top of Adam optimizer improves the accuracy of PINNs by 1-2 orders of magnitude, whereas it improves the accuracy of PISNs and PINSNs by 1 order of magnitude. Additionally, for all output variables of interest in both the examples, we observe the accuracy of PISN improves by an order when the depth of the symbolic network is increased from 2 to 3, whereas there is no substantial improvement on further increasing the depth to 4. This observation further extends to PINSNs, with PINSN of Depth 3 being an order of magnitude better than PINSN of Depth 2. On an average, the best performing PINSNs are 4.71 and 4.98 times better than best-performing PINNs on Kovasznay flow (Navier Stokes) and two-dimensional Burger's equation respectively. 

\end{document}